# Multivariate Gaussian Topic Models: A novel approach to discover topics with high semantic coherence

Satyajeet Sahoo[1], and Jhareswar Maiti[1,2]

[1]Department of Industrial Engineering, Indian Institute of Technology Kharagpur, Kharagpur-721302, West Bengal, India
[2]CoE in Safety Engineering and Analytics, Indian Institute of Technology Kharagpur, Kharagpur-721302, West Bengal, India

**Abstract.** An important aspect of text mining involves information retrieval in form of discovery of semantic themes (topics) from documents using topic modelling. While generative topic models like Latent Dirichlet Allocation (LDA) or Latent Semantic Analysis (LSA) elegantly model topics as probability distributions and are useful in identifying latent topics from large document corpora with minimal supervision, they suffer from difficulty in topic interpretability and reduced performance in shorter texts. Here we propose a novel Multivariate Gaussian Topic Model (MGTM). In this approach topics are presented as Multivariate Gaussian Distributions and documents as Gaussian Mixture Models. Applying EM algorithm on a document corpus, the various constituent Multivariate Gaussian distributions corresponding to the latent topics and their respective parameters are identified. Analysis of the parameters of each distribution helps identify the respective topic keywords, and from these key-words topic annotations are carried out. This approach is applied on 20 newsgroups dataset to demonstrate the interpretability benefits vis-à-vis 4 other benchmark models. The effectiveness of this model in capturing the semantic theme of the topics with high interpretability is examined by calculating the topic coherence and comparing the coherence values with benchmark models. This model achieves a highest mean topic coherence (0.7) and median topic coherence (0.76) vis-à-vis the benchmark models, demonstrating high effectiveness in identifying interpretable, semantically coherent topics.

## 1 Introduction

In the era of internet 4.0, huge volumes of text data are generated worldwide as a consequence of electronic communication and information storage activities. Even offline, text data assumes great importance as part of written language and communications between people, social institutions, and commercial organizations. In organizations, huge volumes of text documents are generated as part of organizational activities and operations containing crucial information pertaining to various stakeholders. Hence it becomes imperative to use text mining methods to discover high quality information, insights, and themes from the text in an automated manner. In supervised text mining, labelled documents are used to train model parameters, which are then applied on test documents to perform classification and prediction. However, when the volume of text documents is huge, it becomes resource intensive and time consuming to manually create and annotate a training dataset. Hence unsupervised models hold great promise to help identify latent themes in text data without the resource and time investment in creating annotated dataset.

One of the important unsupervised text mining activities involves identification of inherent themes/ideas, also called topics, from document corpus using topic modelling. Prominent topic models like Latent Dirichlet Allocation (LDA) and Latent Semantic Analysis (LSA) belong to the class of generative topic models, where for each word position in a document, the model generates a multinomial document distribution of topics for that word position to probabilistically identify the most likely topic corresponding to that position and then generates multinomial distribution of words for that topic to identify the most probable word corresponding to that position. However, these models suffer from drawbacks of poor interpretability/semantic coherence. It has been found that the identification of the topic names from the keywords is very difficult. Secondly, these models assume generation of words at each position as an independent event and does not factor in the influence of other words in generation of that word. Thirdly, LDA can identify multiple topics even for short texts i.e. even for a text consisting of one or two sentences, LDA may identify 2-3 topics which is not very realistic. To address the limitations of these models, various neural topic models like Top2Vec and BERTopic have been proposed, with significant improvement in topic interpretability and coherence.

In this paper, the authors propose an alternate framework of modelling topics from text documents in an unsupervised manner called Multivariate Gaussian Topic Model (MGTM). Instead of considering topics as multinomial distribution of words, topics are presented as Multivariate Gaussian distributions with tokens as variables. Hence the framework considers a document as a multivariate Gaussian Mixture Model (GMM) where the component Multivariate Gaussian distributions represent topics, each topic represented by its unique mean vector and covariance matrix of words covering the vocabulary. The objective is to identify the unique mean vectors and covariance matrices for each Gaussian distribution using Expectation Maximization (EM) algorithm, and then identify the variables (words) that have highest contribution to the mean and covariance values in the mean vector/covariance matrix. These words then become the representative keywords for the topic. The authors demonstrate that the set of keywords obtained using this method achieve the highest coherence values compared to popular and state-of-the-art topic models like LDA, LSA, BERTopic and Top2Vec, signifying high interpretability and effectiveness in obtaining semantically coherent topics.

The remainder of this paper covers the methodology and presents a case study showcasing the application of this model. Section 2 covers the trends in research in domain of topic modelling gaps and challenges in application of traditional topic models. Section 3 briefly describes the concepts behind various models/methods used in this study. Section 4 gives reasoning behind considering topics as Gaussian distributions and the use of EM algorithm. Section 5 describes the sequential steps carried out in implementing the model and obtaining the keywords. In Section 6 the model is applied on a benchmark 20 Newsgroups Dataset, topic keywords are obtained using MGTM as well as LDA, LSA, BERTopic and Top2Vec models and interpretability compared using coherence metric. Finally, Section 7concludes with a summary of the contributions, future scope and future opportunities for model application.

## 2 Literature Review

Text mining is carried out to extract high quality information from textual data with minimal human input [1]. Allahyari et al [2] classified various text mining approaches as Information Retrieval (IR), text summarization, Natural Language Processing (NLP), sentiment analysis, opinion mining and Information Extraction (IE). One important type of information retrieval is topic modelling, wherein the goal is to automatically discover latent themes/subjects of a document. Fu et al [3] identified and investigated eight topic modelling methods. Topic models can be classed as non-probabilistic (discriminative) and probabilistic (generative). Some models like LSA (Latent Semantic Analysis), Factor Analysis and Non-negative Matrix Factorization (NMF) are non-probabilistic linear algebraic models which rely on factorization of document-word matrix. On the other hand, probabilistic Latent Semantic Analysis (PLSA) [4], Latent Dirichlet Allocation (LDA) [5] and Hierarchical Dirichlet processes are generative probabilistic models that assume topics and documents as probability distributions that have their unique parameters/hyperparameters, and by tuning the model hyperparameters, these distributions and the original document is attempted to be generated [6]. PLSA and LDA model documents as probability distribution with topics as attributes and topics as probability distributions with words as attributes, and for each word position in the document tuning the parameters of these distributions provides words with highest likelihood of fitting that position.

While generative topic models have shown great success in modelling latent topics, they suffer from certain drawbacks. Topics are expected to be semantically coherent [7]. While modelling topics as probability distributions is an elegant approach, these generative models do not encode the semantic coherence and any such observation of semantic coherence found in the inferred topic distributions, in some sense, accidental [8]. Since the criteria for identifying keywords representing the topics is probability of occurrence, high probability but low information/noisy/spurious/unrelated words can occur together, adding to reduced topic interpretability. Hence calculating and improving semantic coherence of topics is a critical research problem. Chang et al [9] proposed methods of human evaluation of latent space of topics using intrusion of spurious words and topics to measure semantic coherence of topics. Mimmo et al [10] analyzed ways in which topics can be flawed, proposed topic coherence for automated analysis of assessing topic quality and proposed generalized Polya-urn models to improve topic coherence that incorporates corpora specific word-cooccurrence information. Alokaili et al [11] analyzed various word re-ranking methods and demonstrated word reranking improving topic interpretability.

Secondly, these models try to fill up word positions one at a time, the rich relationships between the words and how occurrence of one word affects occurrence of other words is not explicitly built into the models. Hence these models do not greatly factor in correlations between words and hence, between the topics themselves [12]. Loss of correlation affects topic coherence, since presence of high-probability but low-correlated random words, makes it difficult to fit the words into a larger narrative. Considering this drawback, researchers have tried various approaches use word inter-dependence to improve topic semantic coherence. Andrzejewski et al. [13] used Dirichlet prior over the topic-word multinomials to model topic correlation by using domain knowledge to classify links between words into Must-links and Cannot-links, where the links denote the probability of occurrence of both the words in a topic. Blei and Lafferty [14] proposed a correlated topic model which incorporated correlation between the topics. Petterson et al. [15] considered word information not as constraints but as features and proposed a topic-word prior to classify similar words to similar topic distributions. Newman et al. [16] incorporated word relations by applying quadratic regularizer and convolved Dirichlet regularizer over topic-word distributions. Xie et al [12] proposed MRF-LDA where Markov Random Field representing word correlations were imposed on latent topics from LDA.

Here we are proposing multivariate Gaussian Topic modelling, where topics are modelled as multivariate Gaussian distributions, with their unique mean vectors and covariance matrices incorporating the covariance/correlation among words. Various studies have been carried out in Gaussian Topic Modelling. Agovic and Banerjee [17] proposed Gaussian Process Topic Model (GPTM), that extends correlated topic models [14] by capturing correlations among topics as well as leverage known similarities among documents with the help of a kernel. Das et al. [8] proposed Gaussian LDA, where documents are not represented as sequences of word types but as sequences of word embeddings, and a topic is characterized as a multivariate Gaussian Distribution. The rest of the operations are carried out as LDA. A similar approach was applied by Hu et al [18] to analyze audio, with assumption that an audio recording can be considered as comprising of several topics in form of Gaussian distributions over certain audio attributes.

In our model, we also propose topic as a multivariate Gaussian Model with each word in the vocabulary representing a variable, using TF-IDF word embeddings to represent words as continuous random variables. Also, rather than model the word and topic at the nth word index in a document using LDA, we propose to model all the words representing a topic in a document at once. For this purpose, instead of using

LDA we use EM algorithm to identify the Gaussian distributions with their unique mean vectors and covariance matrices. Analysis of mean vector and covariance matrix for a topic provides the set of all the words having maximum contribution to covariance of a topic, and these become the top words of the topic. Our contributions in this study are as follows: (a) We identify topics by co-occurrence of inter-related words. By modelling topics as Gaussian distributions and analyzing their covariance matrices, we incorporate relations between the words (b) We propose a new metric- covariance explained- to identify topic keywords based on how much covariance they capture (c) Thus in this model instead of finding the most probable words representing the topic, we propose a new metric- Sum-squared mean-covariance contribution (SMCC)- and find the words having highest mean-covariance metric values, which then become the keywords defining the topic. This results in better interpretability vis-à-vis LDA. (d) LDA suffers from lower performance in modelling short texts. LDA sometimes, by its very generative architecture, allots multiple topics to very short texts (consisting of 1-2 sentences). Our topic model allots unique topic to short texts and is able to model short texts with good interpretability.

# 3 Preliminaries

## 3.1 Multivariate Gaussian Distribution

Multivariate Gaussian Probability Distribution (Henceforth referred to as MGD) is generalization of Gaussian Distribution over several dimensions (variables). If there are N variables $v_1, v_2, \ldots v_N$, , then the probability density function (pdf) of an MGD is:

$$\text{Pdf}= P(v, \Theta) = \frac{1}{2\pi^{p/2}|\Sigma|^{1/2}} e^{-(\frac{1}{2})(v-\bar{v})^T\Sigma^{-1}(v-\bar{v})} \quad (1)$$

where Θ stands for the parameter set consisting of

$$\text{mean vector } (\bar{v}) = \begin{bmatrix} \bar{v}_1 \\ \bar{v}_2 \\ \ldots \\ \bar{v}_N \end{bmatrix} \quad (2)$$

which is the vector of individual variable means and

Covariance matrix (Σ) =

$$\Sigma = \begin{bmatrix} var\ (v_1) & cov\ (v_1,\ v_2) & \ldots & cov\ (v_1,\ v_n) \\ cov\ (v_2,\ v_1) & var\ (v_2) & \ldots & cov\ (v_2,\ v_n) \\ \ldots & \ldots & \ldots & \ldots \\ cov\ (v_\text{n},\ v_1) & cov\ (v_\text{n},\ v_2) & \ldots & var\ (v_n) \end{bmatrix} \quad (3)$$

Where $cov\ (v_\text{j},\ v_k)$ = covariance between two variables $v_j$ and $v_k$ and $var\ (v_\text{j})$ = variance of variable $v_j$.

### 3.2 Gaussian Mixture Models

Sometime the distribution of data cannot be explained by a single Gaussian distribution. In such cases a linear combination of multiple Gaussian distributions can better model the data characteristics and distribution. Such linear combinations of two or more basic Gaussian distributions result in composite probability distributions known as mixture distributions. These distributions have three types of parameters: mean vector and covariance matrix of each basic Gaussian, and coefficients of linear combination. By tuning and optimizing the parameters, it is almost possible to approximate any continuous probability distribution to arbitrary accuracy.

This model of representing continuous distributions as a linear combination of basic Gaussians is called Gaussian Mixture Model. Mathematically it is represented as

$$P(v) = \sum_{j=1}^{N} \alpha_j N(v|\mu_j, \Sigma_j) \tag{4}$$

Where $N(v|\mu_j, \Sigma_j)$ is jth basic Gaussian distribution having its own mean vector $\mu_j$ and covariance matrix $\Sigma_j$. The parameters $\alpha_j$ are the weights applied on the Gaussians and are called mixing coefficients. These coefficients are selected such that

$$\sum_{j=1}^{N} \alpha_j = 1 \tag{5}$$

As the probability density functions p(v) and $N(v|\mu_j, \Sigma_j)$ are both greater than 0, hence $\alpha_j \geq 0$ for all j. Substituting in equation (5) we get

$$0 \leq \alpha_j \leq 1 \tag{6}$$

Consider a set of random variables $y_1, y_2 \ldots y_N$ that can take binary values 0 and 1. Let for a particular j, $y_j = 1$ represents selecting the jth component of the Gaussian mixture, and all other y become 0. Hence $\alpha_j = p(y_j = 1)$ represents the marginal distribution of selecting the jth mixing coefficient, and $N(v|\mu_j, \Sigma_j) = p(v|y_j = 1)$ represents the conditional distribution of selecting the jth Gaussian for v. Hence the joint distribution is as represented as

$$p(v) = \sum_y p(y)p(v|y) = \sum_{j=1}^{N} \alpha_j N(v|\mu_j, \Sigma_j) \tag{7}$$

Considering $\alpha_j$ as the prior probability of $y_j = 1$, the posterior probability $p(y_j = 1|v)$, is given by

$$\Lambda\,(y_j) \equiv p(y_j = 1|v) = \frac{\alpha_j N(v|\mu_j, \Sigma_j)}{\sum_{l=1}^{N} \alpha_l N(v|\mu_l, \Sigma_l)} \tag{8}$$

### 3.3 EM Algorithm

If there are K data points, which can be represented by a Gaussian Mixture, assuming the data is i.i.d, the likelihood function is given by

$$P(v|\alpha,\mu,\Sigma) = \prod_{k=1}^{K} \sum_{j=1}^{N} \alpha_j N(v|\mu_j, \Sigma_j) \quad (9)$$

Taking the log-likelihood we get

$$\ln(P(v|\alpha,\mu,\Sigma)) = \sum_{k=1}^{K} \ln\left(\sum_{j=1}^{N} \alpha_j N(v|\mu_j, \Sigma_j)\right) \quad (10)$$

The MLE (Maximum Likelihood Estimate) for the Gaussian Mixture is calculated using Expectation-Maximization (EM) algorithm. The likelihood function in equation (10) is maximized by taking partial derivatives w.r.t parameters (mean vector, covariance matrix and mixing coefficient) and equating the derivatives to 0. Further arranging, we get

$$\mu_j = \frac{1}{K_j} \sum_{k=1}^{K} \Lambda(y_{kj}) v_k \quad (11)$$

$$\Sigma_j = \frac{1}{K_j} \sum_{k=1}^{K} \Lambda(y_{kj})(v_k - \mu_j)(v_k - \mu_j)^T \quad (12)$$

$$\alpha_j = \frac{K_j}{K} \quad (13)$$

EM algorithm involves carrying out the following steps:

1. Assign random initial values to $\mu_j$, $\Sigma_j$ and $\alpha_j$, and calculate initial likelihood estimate.
2. **E step**. Evaluate the posterior probabilities using the parameter values of step 1:

$$\Lambda(y_{kj}) \equiv \frac{\alpha_j N(v|\mu_j,\Sigma_j)}{\sum_{l=1}^{N} \alpha_l N(v|\mu_l,\Sigma_l)} \quad (14)$$

3. **M step**. Parameters' Re-estimation using the calculated posterior probabilities of step 2:

$$\mu_j^{New} = \frac{1}{K_j} \sum_{k=1}^{K} \Lambda(y_{kj}) v_k \quad (15)$$

$$\Sigma_j^{New} = \frac{1}{K_j} \sum_{k=1}^{K} \Lambda(y_{kj}) \left(v_k - \mu_j^{New}\right)\left(v_k - \mu_j^{New}\right)^T \quad (16)$$

$$\alpha_j^{New} = \frac{K_j}{K} \quad (17)$$

4. Evaluate the log likelihood as given in equation (10) and check for convergence. Stop if convergence satisfied else return to step 2.

### 3.4 Coherence

To evaluate topic models, it is important that the topics are well interpretable i.e. they can be well identified from the keywords. A good topic model should give topics where the words are semantically related. One of the metrics used to evaluate the semantic relations between word sets to measure topic interpretability is coherence. Distributional hypothesis states that words that occur in similar contexts represent similar meanings. Hence various word-sets occur together in similar environments. This is the base on which coherence scores are built. One common coherence score used is Cv Coherence, that compares each word in a topic with all topic sets. For all probable words per topic, word vectors are created containing Normalized Pointwise Mutual Information (NPMI) between that word and other words in the vector. Then all word vectors in a topic are combined to form a global topic vector. The average of all cosine similarities between each topic word and its topic vector is used to calculate Cv score. Cv score is calculated by:

$$\mathrm{Cv} = \frac{\sum_{k=1}^{K}\sum_{n=1}^{N} s_{cos}(\vec{w}_{n,k},\vec{w}^{*}{}_{k})}{N\ x\ K} \tag{18}$$

## 4 Why topic is Multivariate Gaussian?

In a probabilistic topic model, the probability of occurrence of a word $w_i$ is given by

$$\mathrm{P(w_i)} = \sum_{j=1}^{N} \mathrm{P}(\Theta_j)\mathrm{P}(\mathrm{w_i}|\Theta_j) \tag{19}$$

Hence for occurrence of k words $\mathrm{w_i} \dots \dots \mathrm{w_K}$ in the document the model is given as

$$\mathrm{P(w_1..w_K)} = \mathrm{P(w_1)P(w_2)..P(w_K)} = \prod_{i=1}^{K}\sum_{j=1}^{N} \mathrm{P}(\Theta_j)\mathrm{P}(\mathrm{w_i}|\Theta_j) \tag{20}$$

This model assumes that the words are independent and identically distributed. However, occurrence of words is not independent or identically distributed; words frequently co-occur, and occurrence of one word influences the occurrence of some other words. E.g. words like “supply” and “chain” frequently co-occur. Hence to develop a topic model, co-occurrence of words has to be factored in and word occurrences cannot be treated as independent events. Hence the model should model the occurrence of every word in the vocabulary W at the same time, given as

$$\mathrm{P(W)} = \sum_{j=1}^{N} \mathrm{P}(\Theta_j)\mathrm{P}(\mathrm{W}|\Theta_j) \tag{21}$$

Then a topic can be considered as a Multivariate Gaussian Distribution that models the occurrence of entire vocabulary vector of words simultaneously as a word vector W

$$\mathrm{W^T} = [w_1 \quad w_2 \quad \dots \quad w_K], \tag{22}$$

With each word representing a random variable.

**Proof:** Considering entire vocabulary vector W (22), equation (21) can be re-written

$$P(W) = \sum_{j=1}^{N} P(\Theta_j) P(W|\Theta_j) \tag{23}$$

Also, since probability distributions can almost be approximately modelled as Gaussian mixtures, probability of occurrence of the word vector in a document, P(W) can be modelled as

$$P(W) = \sum_{i=1}^{N} \alpha_i N(W|\mu_i, \Sigma_i) \tag{24}$$

Since $\alpha_j \geq 0$ and $\sum_{j=1}^{N} \alpha_j = 1$, hence $\alpha_j$ satisfies the necessary conditions of a probability function. Equating (23) and (24), we get

$$\sum_{j=1}^{N} P(\Theta_j) P(W|\Theta_j) = \sum_{j=1}^{N} \alpha_j N(W|\mu_j, \Sigma_j) \tag{25}$$

Now,

$P(\Theta_j)$ = Probability of occurrence of topic $\Theta_j$ and its MLE is given by

$$P(\Theta_j) = \frac{N_i}{N} \tag{26}$$

where $N_j$ is the number of data points (documents) assigned to topic $\Theta_j$ and N is total number of documents.

But, as per EM algorithm, $\alpha_j = \frac{N_j}{N}$

$$\text{Hence, } P(\Theta_j) = \alpha_j \ \ \forall\ i \in \{1, N\} \tag{27}$$

Now, (25) can be reduced to

$$\sum_{j=1}^{N} \frac{N_j}{N} P\ (W|\Theta_j) = \sum_{j=1}^{N} \frac{N_j}{N} N\ (W|\mu_j, \Sigma_j) \tag{28}$$

Since (28) is an identity, it is satisfied when

$$P(W|\Theta_j) = N(W|\mu_i, \Sigma_i)\ \forall\ i \in \{1, N\} \tag{29}$$

Hence a topic can be considered as a multivariate gaussian distribution of vocabulary vector W, where each unique word is a variable. A document can be considered as a Gaussian Mixture Model of topics.

Reasoning behind simultaneous consideration of all words: To answer this, we need to analyze the question:

### 4.1 What is the event in case of writing a topic?

The event is not writing words of topic, one word at a time. The event is putting forth an idea in the form of words constituting a topic in front of a reader. To develop a generative model for topics, we need to emulate topic generation the way the original author generates and writes topics by visualizing how ideas are given a concrete shape and structured in mind. When we think about a topic, some keywords simultaneously appear in our mind shaping the topic. For example, if is we want to write about a gas cylinder accident, keywords like gas, cylinder, kitchen, explosion, fire simultaneously come to the mind. So, a topic on gas cylinder accident is expressed by simultaneous presentation of these keywords on paper. Even if the words appear to have been written one by one (which is because of the writing structure of the language which limits us to write one word at a time), the set of keywords for a topic have already been selected in the mind, and are merely waiting to be put on paper. In probabilistic terms, it is no longer (probability of occurrence of a word in a topic $P(w_i|t)$ for each word position one-by-one sequentially, but rather probability of multiple occurrences of all the unique words in the vocabulary simultaneously $P(w_1, w_2, \ldots w_n|t)$. If the unique words $w_1$, $w_2 \ldots w_n$, each are considered as variables of the event of topic-writing, then we are measuring the random variable for each of the words in the vocabulary simultaneously, thus simulating a multivariate setting for each topic. These values represent the contribution of each variable (word) in expressing the semantic theme of the topic.

If each unique word is a variable, what will be the random variable function? Suppose you have to build a topic. You have sacks of words, with each sack representing a word in the vocabulary. You can pick as many words from a sack as you want in order to build the topic structure. Consider a topic as a collection of pre-determined N words. So, to build the topic we have to pick N words form all the sacks, where some words (potential key words) will be picked multiple times e.g. word 1 may be picked $n_1$ times, word 2 will be picked $n_2$ times and so on, while majority of the non-relevant words will not at all be required, making their value =0. Hence, we will have $n_1 + n_2 + \ldots + n_k = N$. Using the sack analogy, you will pick word 1 occurring $n_1$ words from sack corresponding to word 1, $n_2$ words from sack corresponding to word 2 and so on. times, and so on. This is in contrast to the multinomial assumption of PLSA and LDA models, where for each word position we are essentially throwing a V-sided dice (V being the size of the vocabulary) and selecting the post probable word occurrence among the V occurrences, and repeating this dice throw for each word position, here we select words from the sacks for all word positions at once.

Hence the random variable corresponding to each word variable is the frequency of the word occurrence. However, frequency is a discrete variable, and Gaussian topic model requires continuous variables. To convert the frequency into a continuous variable, in this study we have used the TF-IDF values. As some words that form the crux of the topic will be picked more frequently (exhibiting high TF-IDF values) and other words will be picked sparsely (exhibiting low TF-IDF values), the distribution of TF-IDF values for each contributory word variable for a topic in each document in the corpus is assumed to form a Gaussian bell curve high TF-IDF values for keywords and tapering off to TF-IDF values as we move along sparser words. Then a topic will form a multivariate Gaussian distribution of all words in the corpus, with a mean vector comprising means of all words in the corpus and covariance matrix measuring covariance of each pair of words. The topic j then becomes $N(W|\mu_j, \Sigma_j)$.

Hence analysis of topics requires analysis of the mean vector and covariance matrix. Once important task of topic modelling is topic annotation by finding keywords. Here, models like LDA and LSA consider keywords as most probable words, in this study we utilize the fact that in a topic, critical words co-occur i.e. they present significant covariance and correlation. Since the semantics of a topic is understood when we study the words together with their relations, covariance can be considered as a proxy for semantic theme. Hence those word variables that have highest contribution both in mean and covariance values are taken as potential keywords.

## 5 Methodology

The taxonomy of MGD Topic Model is presented in Figure 1:

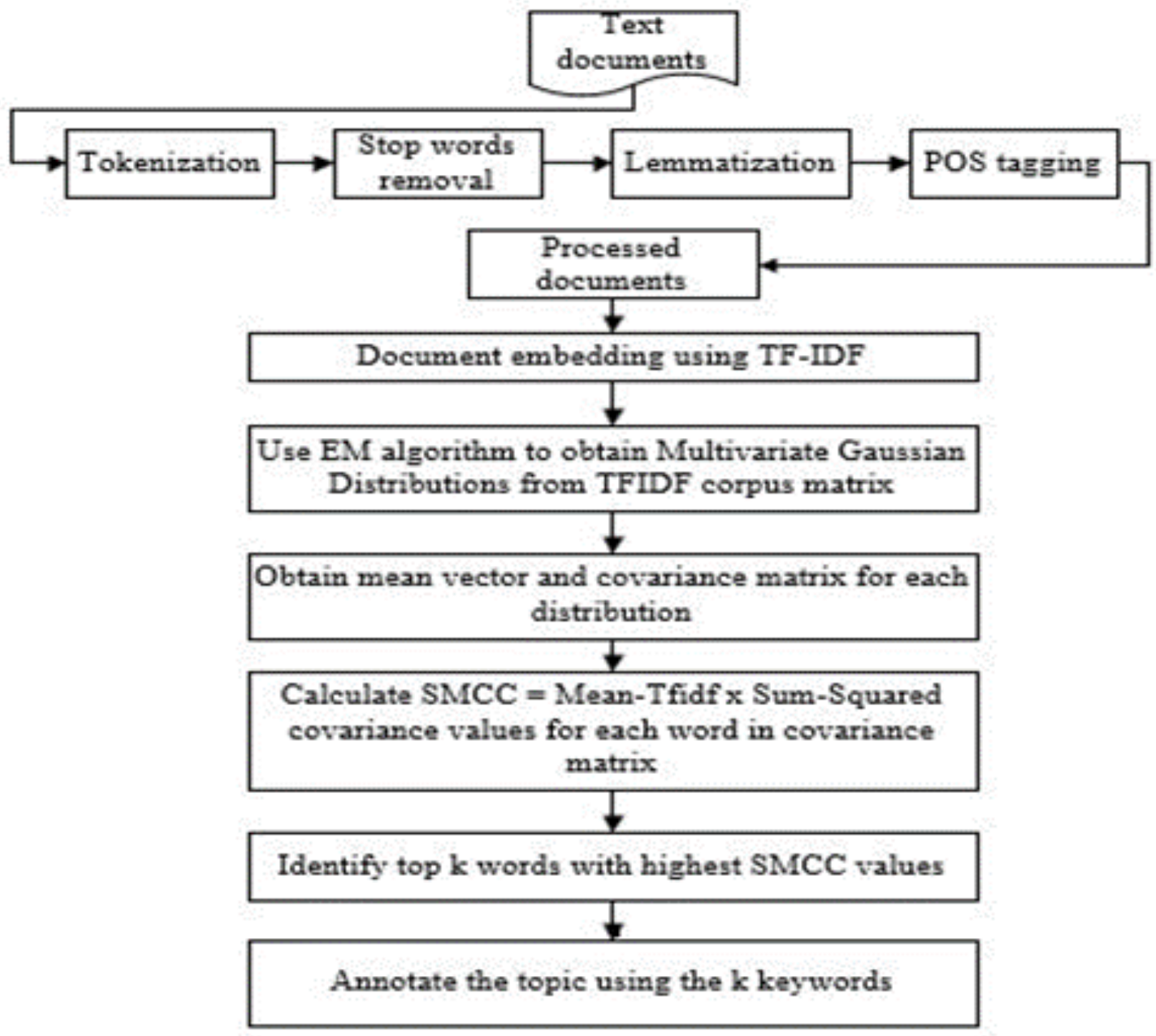

Fig. 1. Flowchart of Multivariate Gaussian Topic Annotation

The various steps for identification of the to topics and associated key-words using MGT Model are:

### 5.1 Text Preprocessing

The first step involves standardizing the unstructured, free-flowing text into a structured form and identifying the words as random variables. This is called pre-processing of text and involves the following steps:

*Lower-case conversion and removal of punctuations:* Presence of punctuations and characters in lower-case and upper-case adds complexity to the data. Hence characters are uniformly converted to lower case. Similarly, punctuations are removed to reduce complexity without incurring significant information loss.

*Removal of Stop-words:* Some words in text are used as fillers and hence they have high frequency of occurrence e.g. prepositions and articles like “for”, “the”, etc. Such words and other common words, referred to as stop-words, have low information value in the text. Hence these are removed during pre-processing [19] [25]. For this purpose, either the Snowball Stemmer [20] package or the Terrier package [21] is used as they contain lists of stop-words.

*Tokenization:* Tokens are the building blocks of text. Intext pre-processing the individual words are considered as building blocks and are characterized as tokens and separated using tokenization [25]. These tokens are separated from running text using delimiters like newline, tab, space etc. [22].

*Lemmatization:* There are various morphological variations of words in a text e.g. ‘injury’, ‘injurious’ etc. These word variations are standardized, either by removing word endings (stemming), or conversion to a base form called lemma (lemmatization) e.g. ‘bank’, ‘banker’ ‘banking’ become ‘bank’ using lemmatization. [23]. In this study, stemming approach is not used since removing word endings makes then difficult to interpret, hence only lemmatization is used.

*Tagging POS:* In the next step, the tokens are assigned labels as noun, verb, adverb, adjective etc. (Part of speech labels). The list of POS tags from Penn Treebank are used for this purpose [24] [25].

*Word embedding:* In the final pre-processing step, the lemmatized words are embedded i.e. the words are represented by vectors to facilitate application of quantitative models. Since Gaussian distributions require continuous data, and since each word is considered as a variable for the Multivariate Gaussian topic model, hence TF-IDF is used for representing the words with continuous values.

### 5.2 Using EM algorithm to identify gaussian distributions

The preprocessed data was stored in a N x V matrix, where N is the number of documents and V is the corpus vocabulary size. So, each row represents a document consisting of all the pre-processed unique words in the corpus, with their respective TF-IDF values. On the matrix, EM algorithm is applied and the constituent Gaussian distributions are identified, and their respective parameters- mean vectors and covariance matrices- are obtained.

### 5.3 Identification of keywords using Mean-Squared Covariance Metric

In popular probabilistic models like PLSA or LDA, the keywords are determined based on their probability of occurrence in a topic. However, since topics consist of keywords that co-occur, hence the keywords exhibit substantial covariance with each other. As per distributional hypothesis, when two words belong to a topic, their probability of occurrence is high (as they belong to the same distribution) and will exhibit higher covariance. In comparison, two words that belong to different topics will exhibit random co-occurrence (as they belong to different distributions) and will have zero correlation/covariance. Hence words that capture/explain the maximum covariance of the topic are the words that are non-random and co-vary, making their presence in the topic foundational and not coincidental.

Hence, we propose a new metric to identify the keywords of a topic: **Covariance Explained.** Based on this metric, words that exhibit highest covariance with other words should become the keywords of the topic. While variance explained metric is widely used in machine learning models, this proposed Covariance Explained metric is more robust because of the following reasons:

a) Variance is nothing but the covariance of a variable with itself. Hence variance explained model is actually covariance explained model on a broader sense;
b) Variance explained only explains the variance without touching upon the relationship between the variables. Covariance explained explains both the variability of the features as well as the variability based on the relationship among the features.

Once the respective means and the covariance matrix of each of the Gaussian distribution are identified using EM Algorithm, word variables that have greatest contribution to the covariance of the topic distribution are calculated. The total covariance of the corpus is represented by the global V x V covariance matrix, which incorporates the variance of each word variable as well as the covariance values of each pair of words $w_i$ and $w_j$. The covariance contribution of each word is calculated by calculating the sum of squares all the variance and covariance values of the corresponding row in the covariance matrix. Squaring is necessary because even words with negative covariance show relation, albeit inverse, which makes them likely to be related to the topic. Without squaring of covariances, summing of positive and negative covariances may cancel each other out, leading to a low sum which may risk a high-contributing word getting missed out. Once the sum square values are calculated, the words with the highest sum squared values become the potential keywords for the topic.

However, the topic is a Gaussian distribution having two parameters mean and covariance matrix, and both contribute to the semantic theme of a topic. Hence a word that has high values both in the mean vector and sum squared covariance, should be the ideal keyword. Hence, we propose a new metric: **Sum-Squared Mean-Covariance contribution (SMCC)** metric where the sum of squares of variances and covariances are obtained for a word are then multiplied with the respective TF-IDF values from the mean vector to obtain the metric:

$$\text{SMCC}\ (x_i) = \text{TF-IDF}\ (x_i)\ \text{x}\ \sum_{j=1}^{V} \sigma_{ij}{}^2 \tag{30}$$

For each topic, from the mean and covariance matrix identified are used to find the SMCC values for each word and then the words are sorted in descending order of the SMCC values. The top 10 words thus identified become the keywords for that topic.

# 6 Results and Discussion

To test the performance of the MGT model in a generalized setting, benchmark 20-Newsgroups dataset was chosen. This dataset consists of almost 18000 news posts belonging to 20 previously identified topics. MGT Model was applied on this dataset and the 20 topics were identified and the topic keywords were identified using SMCC metric. Also, the performance of MGT model was compared with benchmark topic models-Latent Semantic Analysis (LSA), Latent Dirichlet Allocation (LDA), BERTopic and Top2Vec, which were also applied on the Newsgroups dataset and the topic keywords of the 20 topics were respectively identified. The performance was measured by degree of capture of the semantics of the topic by the keywords, by measuring the coherence metric for each of the models.

Table 1 shows how the keywords are identified for topic 0. For each word i its mean (denoted by $\boldsymbol{\mu}_{0i}$) and the sum squared of covariance (denoted by $\left(\sum_{j=1}^{V} \sigma_{0ij}\right)^2$) were identified and multiplied to get SMCC values. The SMCC values were sorted in descending order and the top 10 words identified were taken as the keywords for the topic. The keywords of rest 19 topics using SMCC is shown in Appendix.

Table 1: Identification of top 10 keywords for Topic 0

| Words | $\left(\sum_{j=1}^{V} \sigma_{0ij}\right)^2$ | $\mu_{0j}$ | $\mu_0 * \left(\sum_{j=1}^{k} \sigma_{0ij}\right)^2$ |
|---|---|---|---|
| mail | 0.00021593 | 0.09879979 | 2.13339E-05 |
| address | 0.00028967 | 0.05269607 | 1.52646E-05 |
| please | 0.00010308 | 0.1098505 | 1.13237E-05 |
| thanks | 0.00011701 | 0.08359365 | 9.781E-06 |
| email | 0.00015472 | 0.05470918 | 8.46442E-06 |
| list | 0.00011687 | 0.03012107 | 3.52032E-06 |
| advance | 7.9915E-05 | 0.04027645 | 3.21871E-06 |
| reply | 7.2312E-05 | 0.03545513 | 2.56382E-06 |
| mailing | 8.6254E-05 | 0.02144531 | 1.84975E-06 |
| send | 4.5098E-05 | 0.03098886 | 1.39753E-06 |

Using the SMCC metric, the identification of the top-10 keywords for each of the topics using SMCC scores for Multivariate Gaussian Topic Model is given in Table 2.

Also, keywords for LSA, LDA BERTopic and Top2Vec are identified, as shown in Tables 3,4,5, and 6:

Table 2: Keywords of MGD Topics

| Topics | Topic Keywords |
|---|---|
| Topic0 | 'mail','address','please','thanks','email','list','advance','reply','maling','send' |
| Topic1 | 'car','engine','oil','dealer','bike','mile','light','price','model','speed' |
| Topic2 | 'god','jesus','christian','church','faith','bible','christ','belief','truth','atheist' |
| Topic3 | 'gun','people','law','weapon','government','right','crime','firearm','would','fbi' |
| Topic4 | 'card','driver','video','bus','monitor','color','vga','slot','diamond','vesa' |
| Topic5 | 'player','year','hitter','hit','pitcher','last','baseball','season','ball','average' |
| Topic6 | 'armenian','greek','turkish','armenia','turkey','turk','genocide','plane','muslim','russian' |
| Topic7 | 'msg','objective','morality','food','moral','value','frank','keith','system','natural' |
| Topic8 | 'edu','uiuc','cso','apr','writes','article','cmu','andrew','news','rutgers' |
| Topic9 | 'com','netcom','article','bike','apr','att','tek','writes','ibm','edu' |
| Topic10 | 'israel','israeli','jew','arab','palestinian','lebanon','soldier','civilian','jewish','cpr' |
| Topic11 | 'sale','shipping','offer','price','condition','asking','manual','excellent','new','disk' |
| Topic12 | 'file','format','window','image','program','ftp','directory','gif','do','zip' |
| Topic13 | 'anyone','thanks','know','appreciated','chip','would','help','appreciate','advance','information' |
| Topic14 | 'key','encryption','chip','clipper','escrow','nsa','phone','bit','government','de' |
| Topic15 | 'space','problem','nasa','one','would','bike','system','max','get','use' |
| Topic16 | 'window','do','font','program','version','manager','application','server','running','event' |
| Topic17 | 'game','team','player','hockey','season','play','nhl','win','leaf','playoff' |
| Topic18 | 'weight','pitt','dsl','cadre','shameful','jxp','chastity','intellect','skepticism','surrender' |
| Topic19 | 'scsi','drive','disk','ide','controller','floppy','boot','hard','tape','mac' |

Table 3: Keywords of LSA Topics

| Topics | Topic Keywords |
|---|---|
| Topic0 | 'edu', 'would', 'writes', 'one', 'com', 'article', 'like', 'know', 'people', 'get' |
| Topic1 | 'window', 'file', 'card', 'thanks', 'drive', 'do', 'driver', 'program', 'system', 'disk' |
| Topic2 | 'edu', 'game', 'writes', 'article', 'team', 'com', 'apr', 'player', 'year', 'season' |
| Topic3 | 'edu', 'com', 'article', 'writes', 'window', 'apr', 'file', 'god', 'do', 'uiuc' |
| Topic4 | 'window', 'game', 'god', 'team', 'file', 'player', 'win', 'jesus', 'christian', 'season' |
| Topic5 | 'drive', 'god', 'scsi', 'card', 'edu', 'disk', 'controller', 'ide', 'hard', 'christian' |
| Topic6 | 'window', 'com', 'car', 'drive', 'driver', 'gun', 'problem', 'do', 'armenian', 'bike' |
| Topic7 | 'car', 'thanks', 'please', 'anyone', 'bike', 'mail', 'looking', 'know', 'advance', 'email' |
| Topic8 | 'com', 'god', 'key', 'chip', 'game', 'netcom', 'jesus', 'card', 'christian', 'bible' |
| Topic9 | 'card', 'driver', 'video', 'israel', 'com', 'israeli', 'armenian', 'monitor', 'arab', 'jew' |
| Topic10 | 'com', 'drive', 'file', 'please', 'disk', 'thanks', 'mail', 'scsi', 'address', 'armenian' |
| Topic11 | 'car', 'window', 'key', 'please', 'israel', 'thanks', 'drive', 'israeli', 'game', 'jew' |
| Topic12 | 'gun', 'would', 'thanks', 'anyone', 'know', 'drive', 'people', 'window', 'advance', 'weapon' |
| Topic13 | 'file', 'card', 'driver', 'car', 'gun', 'ftp', 'format', 'directory', 'site', 'video' |
| Topic14 | 'israel', 'israeli', 'arab', 'jew', 'anyone', 'would', 'bike', 'know', 'palestinian', 'problem' |
| Topic15 | 'armenian', 'turkish', 'armenia', 'know', 'anyone', 'key', 'turkey', 'turk', 'genocide', 'thanks' |
| Topic16 | 'space', 'nasa', 'driver', 'gov', 'car', 'drive', 'card', 'window', 'orbit', 'launch' |
| Topic17 | 'pitt', 'gordon', 'bank', 'geb', 'jxp', 'chastity', 'cadre', 'dsl', 'skepticism', 'intellect' |
| Topic18 | 'bike', 'card', 'motorcycle', 'ride', 'new', 'file', 'rider', 'right', 'dod', 'space' |
| Topic19 | 'would', 'file', 'like', 'price', 'do', 'people', 'chip', 'mac', 'system', 'government' |

Table 4: Keywords of LDA Topics

| Topics | Topic Keywords |
|---|---|
| Topic0 | 'ide', 'scsi', 'drive', 'atheist', 'controller', 'atheism', 'isa', 'peter', 'stage', 'sec' |
| Topic1 | 'edu', 'writes', 'com', 'article', 'would', 'one', 'people', 'god', 'think', 'say' |
| Topic2 | 'april', 'russia', 'canada', 'rock', 'republic', 'france', 'germany', 'usa', 'fool', 'joke' |
| Topic3 | 'address', 'mouse', 'mail', 'mailing', 'list', 'com', 'phone', 'right', 'internet', 'everywhere', |
| Topic4 | 'bos', 'team', 'game', 'cal', 'win', 'brian', 'software', 'season', 'edu', 'vote' |
| Topic5 | 'window', 'adapter', 'reply', 'news', 'moment', 'position', 'electric', 'memory', 'amp', 'manager' |
| Topic6 | 'god', 'marriage', 'gay', 'heart', 'life', 'men', 'hell', 'hole', 'edu', 'eric' |
| Topic7 | 'funding', 'movie', 'ram', 'yes', 'machine', 'minute', 'card', 'vlb', 'diamond', 'color' |
| Topic8 | 'radar', 'helmet', 'size', 'honda', 'ride', 'testing', 'bike', 'receiver', 'brand', 'replace' |
| Topic9 | 'uiuc', 'cso', 'electric', 'edu', 'news', 'writes', 'article', 'cold', 'eat', 'food' |
| Topic10 | 'cub', 'room', 'sent', 'make', 'near', 'move', 'mike', 'today', 'edu', 'ken' |
| Topic11 | 'max', 'part', 'end', 'hey', 'sol', 'fax', 'map', 'vax', 'gon', 'sit' |
| Topic12 | 'billion', 'people', 'poor', 'food', 'group', 'eating', 'cup', 'whether', 'cultural', 'study' |
| Topic13 | 'game', 'devil', 'final', 'warrant', 'jet', 'bitnet', 'lake', 'minnesota', 'writes', 'detroit' |
| Topic14 | 'window', 'thanks', 'edu', 'file', 'anyone', 'drive', 'know', 'card', 'would', 'use' |
| Topic15 | 'douglas', 'bus', 'acns', 'recommend', 'craig', 'classified', 'still', 'measure', 'protect', 'note' |
| Topic16 | 'air', 'automatic', 'valve', 'rear', 'transmission', 'speed', 'tube', 'problem', 'netcom', 'fan' |
| Topic17 | 'game', 'team', 'player', 'edu', 'year', 'season', 'play', 'baseball', 'hockey', 'writes' |
| Topic18 | 'henry', 'film', 'toronto', 'chip', 'com', 'gateway', 'law', 'would', 'imagine', 'bear' |
| Topic19 | 'generated', 'reader', 'shall', 'news', 'probably', 'article', 'solar', 'tree', 'saved', 'salvation' |

Table 5: Keywords of BERTopic Topics

| Topics | Topic Keywords |
|---|---|
| Topic0 | 'edu', 'one', 'writes', 'com', 'would', 'article', 'know', 'people', 'file', 'like' |
| Topic1 | 'people', 'would', 'one', 'key', 'writes', 'edu', 'article', 'right', 'government', 'gun' |
| Topic2 | 'game', 'team', 'player', 'year', 'season', 'play', 'hockey', 'edu', 'win', 'league' |
| Topic3 | 'space', 'car', 'nasa', 'launch', 'would', 'writes', 'satellite', 'orbit', 'article', 'edu' |
| Topic4 | 'window', 'server', 'application', 'widget', 'program', 'file', 'graphic', 'use', 'manager', 'motif' |
| Topic5 | 'health', 'food', 'msg', 'patient', 'doctor', 'disease', 'medical', 'edu', 'article', 'one' |
| Topic6 | 'modem', 'sound', 'card', 'port', 'audio', 'com', 'one', 'offer', 'use', 'please' |
| Topic7 | 'drive', 'scsi', 'disk', 'controller', 'ide', 'hard', 'system', 'speed', 'mhz', 'memory' |
| Topic8 | 'bike', 'car', 'com', 'radar', 'dog', 'helmet', 'writes', 'article', 'detector', 'one' |
| Topic9 | 'card', 'monitor', 'color', 'driver', 'video', 'window', 'vga', 'screen', 'problem', 'bit' |
| Topic10 | 'space', 'edu', 'mail', 'list', 'nist', 'com', 'post', 'address', 'sci', 'article' |
| Topic11 | 'printer', 'font', 'print', 'ink', 'laser', 'deskjet', 'window', 'bubblejet', 'postscript', 'toner' |
| Topic12 | 'sleeve', 'picture', 'sale', 'cd', 'hotel', 'offer', 'promo', 'ticket', 'package', 'room' |
| Topic13 | 'mouse', 'keyboard', 'typing', 'rsi', 'window', 'apple', 'key', 'edu', 'support', 'driver' |
| Topic14 | 'point', 'polygon', 'edge', 'line', 'sphere', 'algorithm', 'plane', 'xxxx', 'problem', 'circle' |
| Topic15 | 'woof', 'say', 'headline', 'kidding', 'art', 'stupid', 'title', 'wanted', 'need', 'many' |
| Topic16 | 'number', 'phone', 'line', 'umass', 'dial', 'ec', 'alee', 'operator', 'call', 'greeting' |
| Topic17 | 'max', 'bhj', 'giz', 'bxn', 'qax', 'chz', 'rlk', 'okz', 'fyn', 'fij' |
| Topic18 | 'stephanopoulos', 'president', 'job', 'going', 'myers', 'think', 'package', 'know', 'secretary', 'administration' |
| Topic19 | 'wire', 'ground', 'outlet', 'wiring', 'neutral', 'circuit', 'conductor', 'grounding', 'gfci', 'breaker' |

Table 6: Keywords of Top2Vec Topics

| **Topics** | **Topic Keywords** |
|---|---|
| Topic0 | 'edu', 'article', 'agreement', 'plaintext', 'consensus', 'upenn', 'grant', 'academic', 'acc', 'norris' |
| Topic1 | 'atheist', 'theist', 'theology', 'believer', 'atheism', 'contradict', 'christian', 'giz', 'fallacy', 'religious' |
| Topic2 | 'honda', 'motorcycle', 'bmw', 'bike', 'automobile', 'toyota', 'drive', 'driving', 'car', 'motor' |
| Topic3 | 'irq', 'scsi', 'cmos', 'dma', 'mhz', 'maxtor', 'motherboard', 'powerbook', 'powerpc', 'processor' |
| Topic4 | 'batf', 'nra', 'firearm', 'idiot', 'atf', 'fbi', 'armed', 'gun', 'guerilla', 'gov' |
| Topic5 | 'myth', 'worst', 'hitler', 'immoral', 'atheism', 'hatred', 'least', 'said', 'doubt', 'apology' |
| Topic6 | 'nyi', 'shark', 'lindros', 'recchi', 'nearly', 'almost', 'said', 'mistaken', 'slower', 'pitt' |
| Topic7 | 'jpl', 'spacecraft', 'nasa', 'orbiter', 'propulsion', 'astronaut', 'satellite', 'orbital', 'planetary', 'ssto' |
| Topic8 | 'recchi', 'nyi', 'nyr', 'gainey', 'potvin', 'nhl', 'selanne', 'lemieux', 'defenseman', 'ahl' |
| Topic9 | 'tiff', 'shareware', 'bmp', 'jpeg', 'bitmap', 'postscript', 'file', 'program', 'printer', 'borland' |
| Topic10 | 'mets', 'clemens', 'phillies', 'pitching', 'alomar', 'recchi', 'inning', 'baseball', 'gainey', 'rbi' |
| Topic11 | 'severe', 'disorder', 'convinced', 'symptom', 'affected', 'studied', 'intervention', 'skepticism', 'substance', 'patient' |
| Topic12 | 'xterm', 'bitmap', 'gui', 'gnu', 'widget', 'printf', 'fprintf', 'borland', 'postscript', 'obfuscated' |
| Topic13 | 'circuit', 'khz', 'analog', 'wiring', 'electronics', 'capacitor', 'amp', 'resistor', 'electronic', 'grounding' |
| Topic14 | 'mailing', 'email', 'mailer', 'mail', 'sending', 'replied', 'reply', 'responded', 'answered', 'thanx' |
| Topic15 | 'obo', 'selling', 'sale', 'powerbook', 'buyer', 'sell', 'gcx', 'trade', 'interested', 'shareware' |
| Topic16 | 'unix', 'msdos', 'bsd', 'linux', 'xterm', 'gnu', 'novell', 'exe', 'irq', 'shareware' |
| Topic17 | 'cryptography', 'encryption', 'encrypted', 'decrypt', 'cryptographic', 'pgp', 'plaintext', 'cipher', 'ciphertext', 'obfuscated' |
| Topic18 | 'ati', 'vga', 'monitor', 'irq', 'motherboard', 'vesa', 'vram', 'svga', 'driver', 'scsi' |
| Topic19 | zionist', 'zionism', 'palestinian', 'israel', 'gaza', 'kurd', 'palestine', 'israeli', 'jew', 'holocaust' |

To measure how MGT topic model performs against other topic model, we try to see how semantically coherent output is produced by each topic model. This is because a topic model should provide keywords that should be coherent enough to identify the topic. For this purpose, the coherence score of each of the 5 topic models is calculated and compared as shown in the figures below:

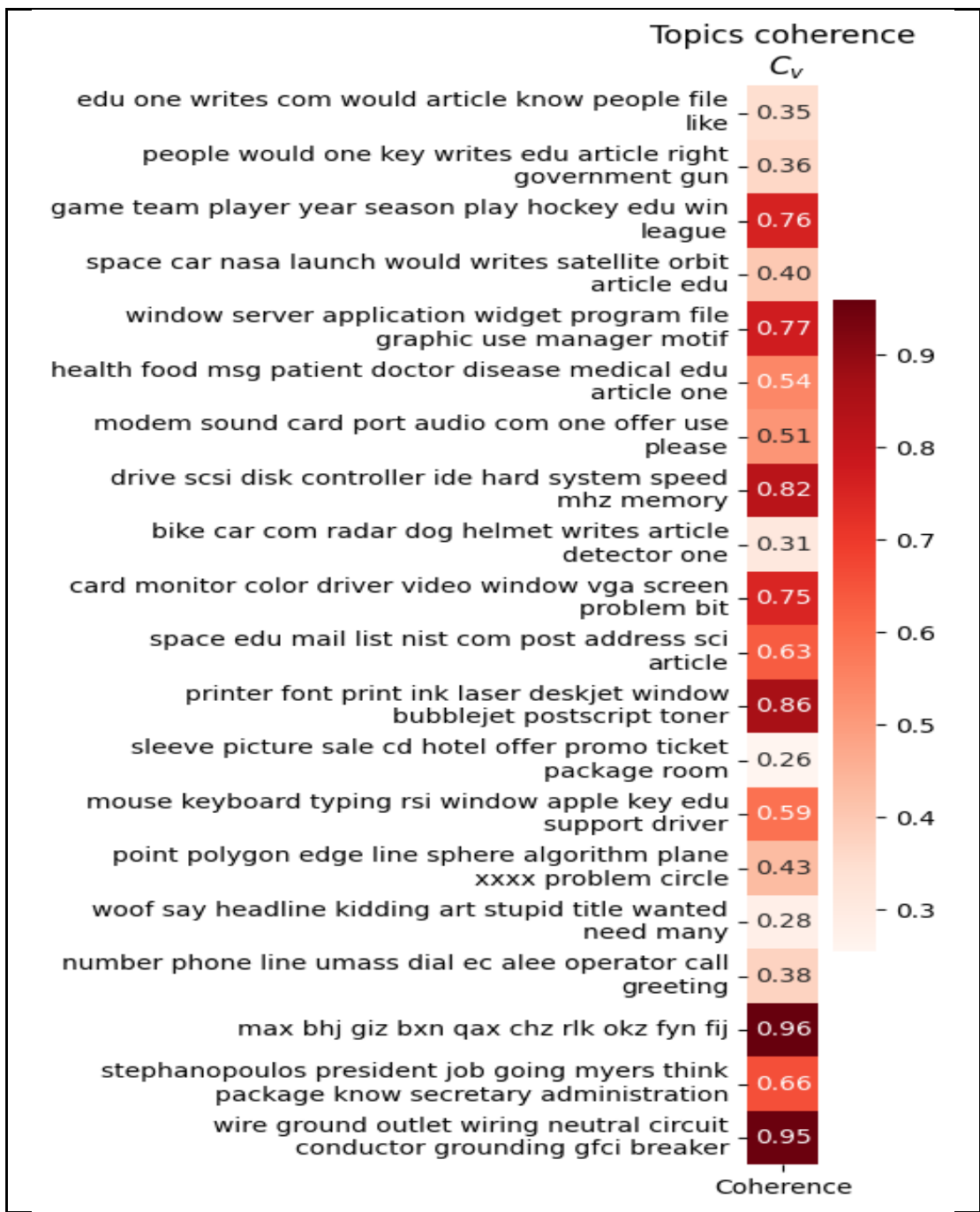

Fig 1: Coherence scores for 20 topics LSA

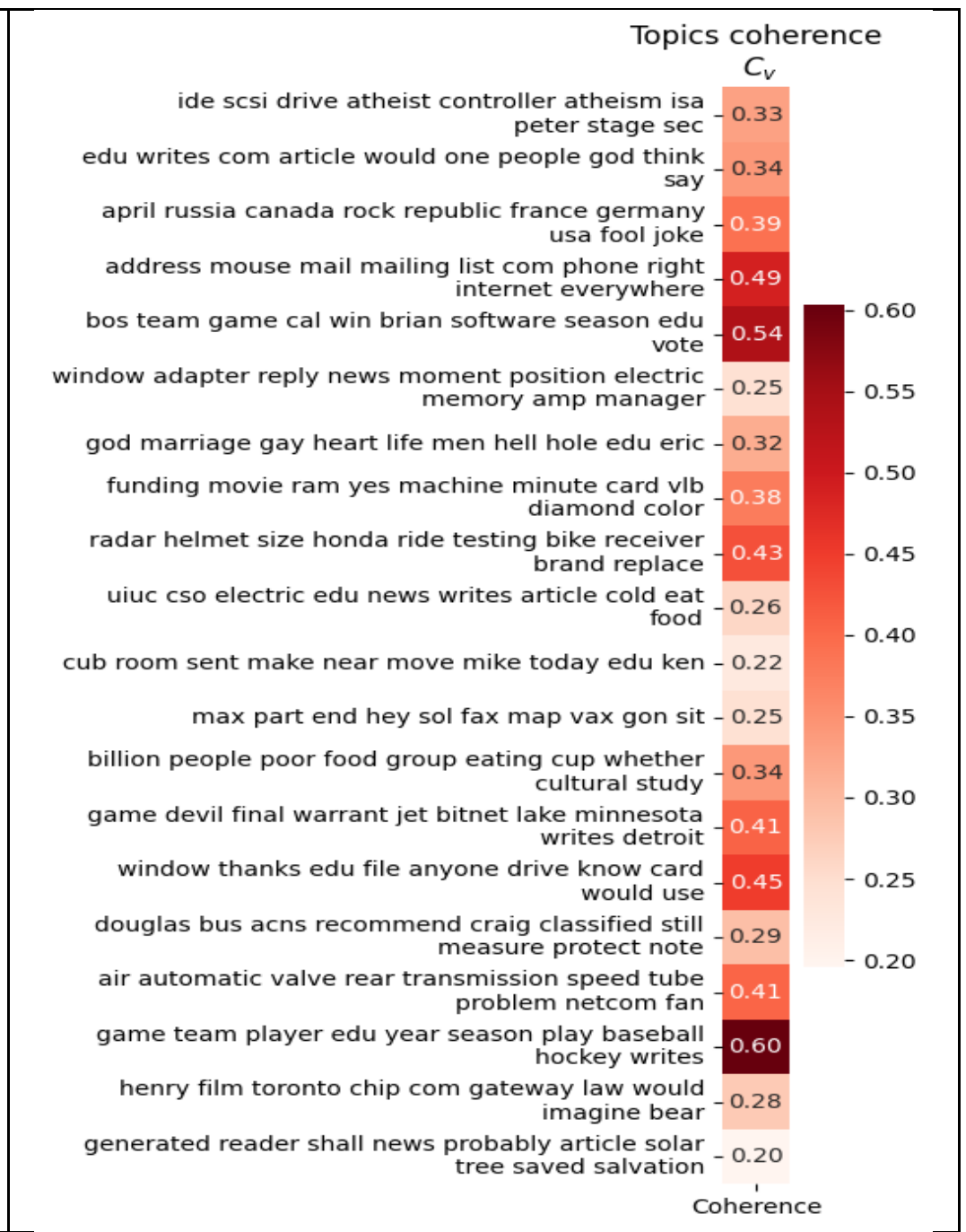

Fig 2: Coherence scores for 20 topics LDA

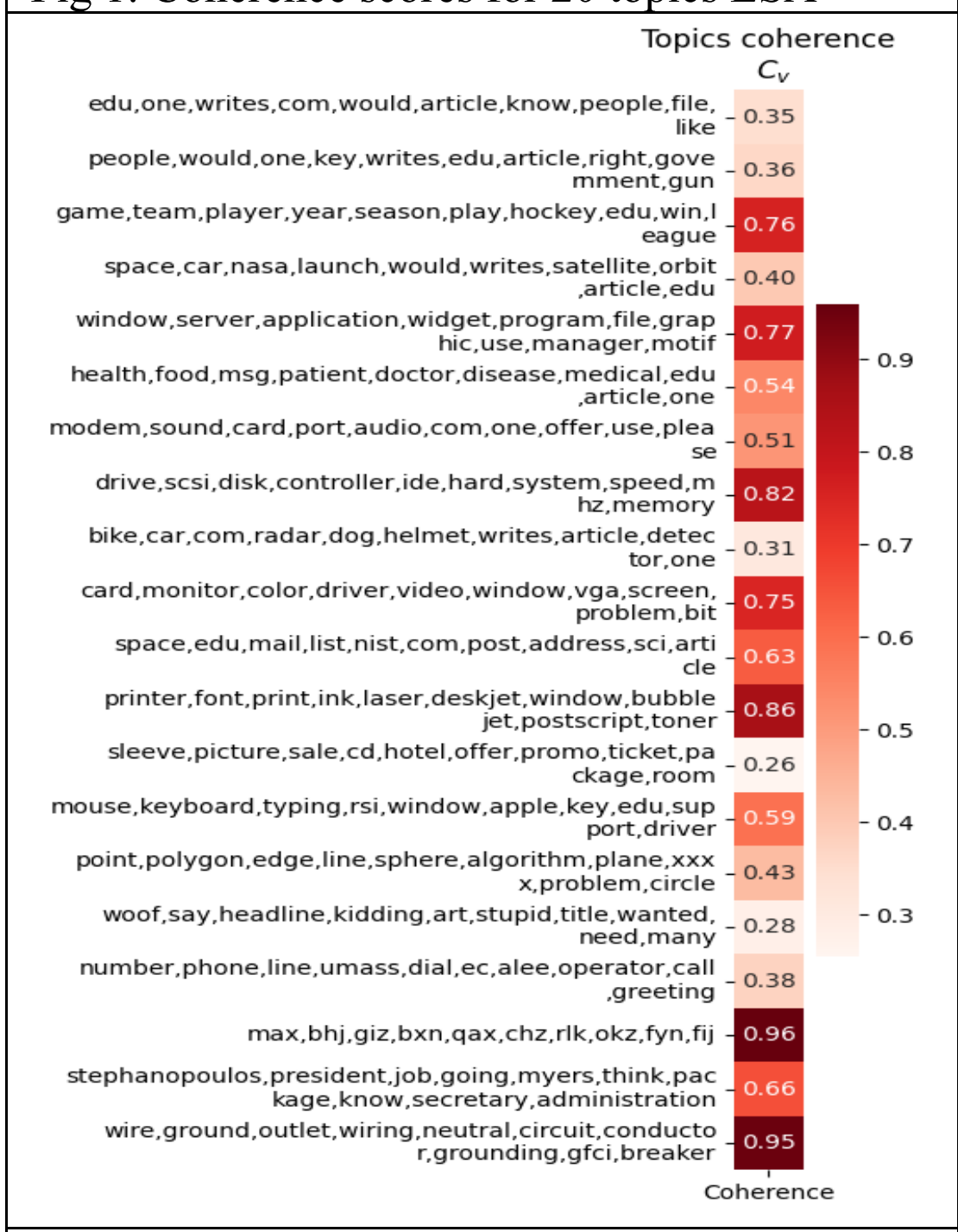

Fig 3: Coherence scores 20 topics BERTopic

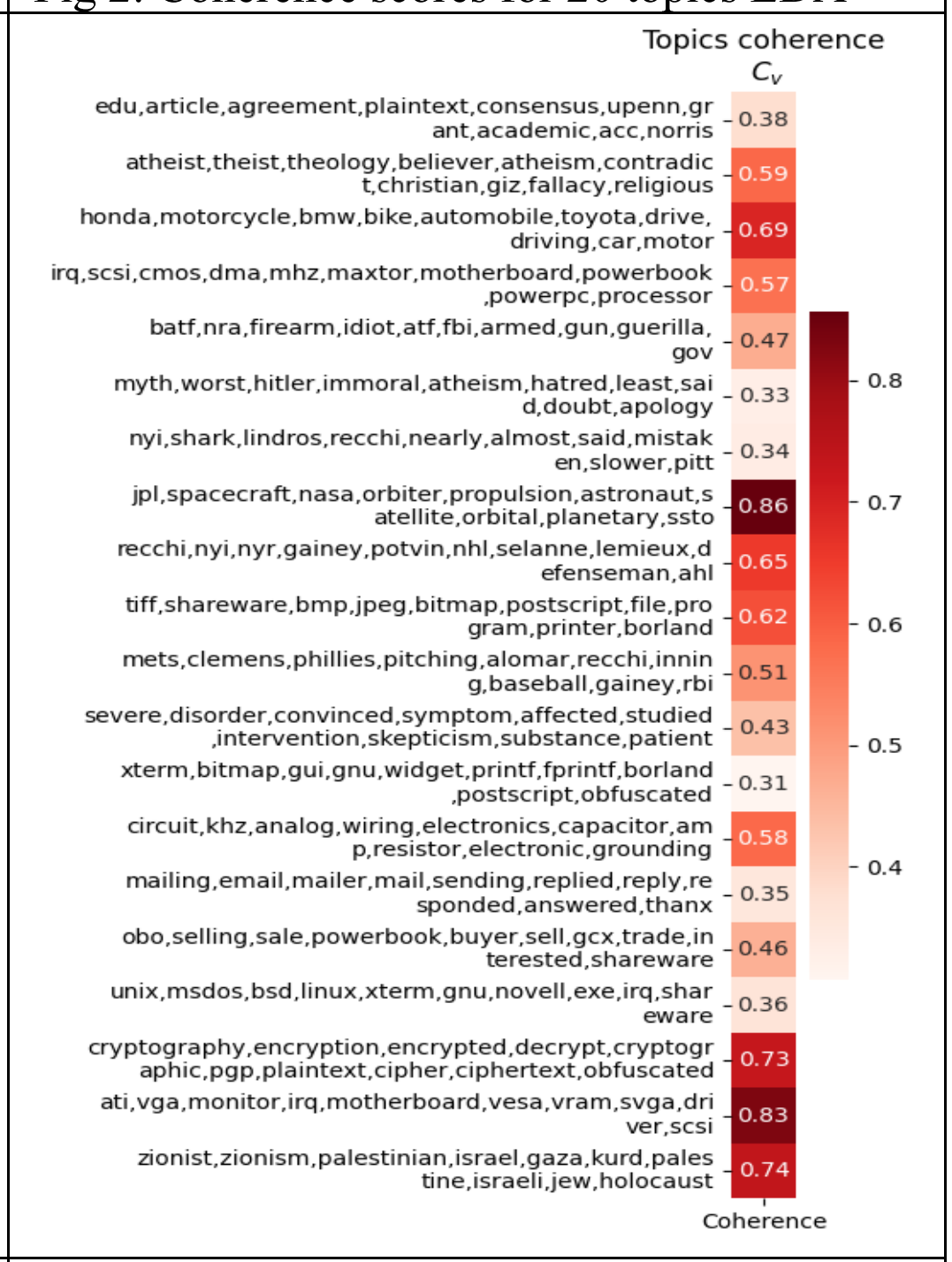

Fig 4: Coherence scores 20 topics Top2Vec

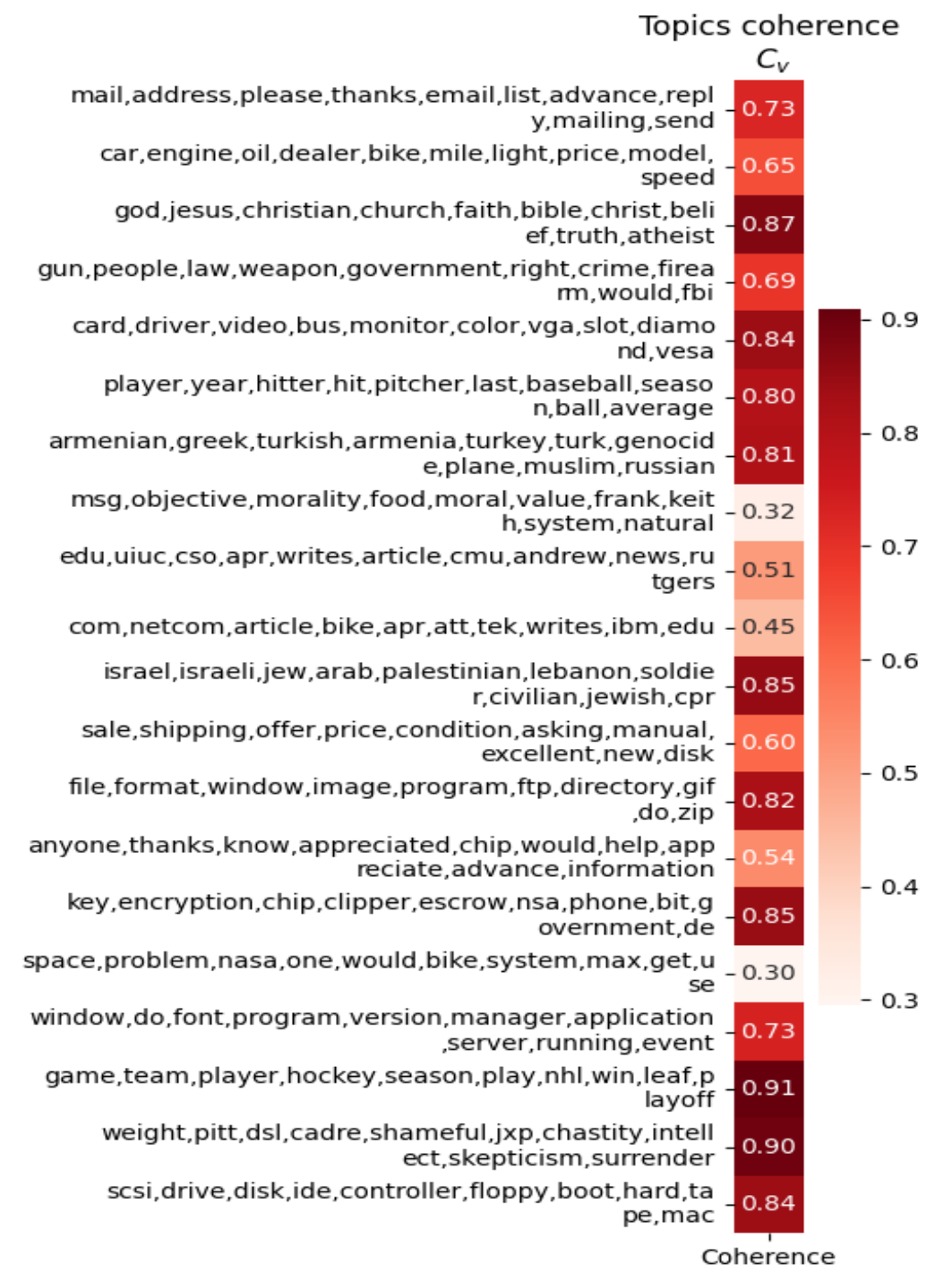

Fig 5: Coherence scores top 20 topics MGT

A comparison of the coherence scores of the 5 topic models on the 20 newsgroups dataset is given in Table 6:

| Metric | LSA | LDA | BERTopic | Top2vec | MGTM |
|---|---|---|---|---|---|
| **Mean coherence** | 0.448 | 0.358 | 0.579 | 0.540 | **0.701** |
| **Median coherence** | 0.433 | 0.341 | 0.568 | 0.541 | **0.768** |
| **Standard Deviation coherence** | 0.161 | 0.105 | 0.224 | 0.167 | 0.185 |

From the above figure, it can be seen that MGT topic model is giving higher mean and median coherence compared to the benchmark topic models.

To find out how effective MGT is in identifying coherent topics, we calculated how many topics were identified by each of the models above a benchmark coherence. The benchmarks were chosen starting from 90% coherence up to 60% coherence. The number of topics identified by each topic model above the respective benchmark is given in Table 7:

| Benchmark | LSA | LDA | MGT | BERTopic | Top2Vec |
|---|---|---|---|---|---|
| **>90% coherence** | 1 | 0 | **2** | **2** | 0 |
| **> 80% coherence** | 1 | 0 | **10** | 4 | 2 |
| **>70% coherence** | 2 | 0 | **12** | 7 | 4 |
| **>60% coherence** | 2 | 1 | **15** | 9 | 7 |

From the table it is evident that MGT model is consistently identifying topics with high semantic coherence in every benchmark.

To statistically prove that MGT is giving higher coherence compared to the benchmarks, we use hypothesis testing.

We consider the following hypotheses:
$H_0$: $\mu_{MGT}$ - $\mu_{Benchmark} = 0$
$H_1$: $\mu_{MGT} - \mu_{Benchmark} > 0$

Where $\mu_{MGT}$ and $\mu_{Benchmark}$ stand for the mean global coherence for MGT and benchmark topic models respectively.

Using t statistic for sample variance (with $\mu_{MGT} - \mu_{Benchmark}$=0 for null hypothesis)

$$t= \frac{(\bar{x}_{MGT}-\bar{x}_{Benchmark})-(\mu_{MGT}-\mu_{Benchmark})}{\sqrt{\frac{s_{MGT}^2(n_{MGT}-1)+s_{Benchmark}^2(n_{Benchmark}-1)}{n_{MGT}+n_{Benchmark}-2}}\sqrt{\frac{1}{n_{MGT}}+\frac{1}{n_{Benchmark}}}}$$

and df = $n_{MGD} + n_{LDA}$ -2, the following results are obtained:

Table 7: t-estimates for the difference of two models

| MGT vs Bench-mark | t-Value (Estimated) | Df | t-Value (Critical) | P-Value | Verdict (Accept/reject Null Hypothesis) |
|---|---|---|---|---|---|
| MGT vs LSA | 4.61 | 38 | 1.686 | 2.2E-05 | Reject |
| MGT vs LDA | 7.21 | 38 | 1.686 | 6.38E-09 | Reject |
| MGT vs BERTopic | 1.88 | 38 | 1.686 | 0.034 | Reject |
| MGT vs Top2Vec | 2.89 | 38 | 1.686 | 0.0032 | Reject |

As the estimated t-value is greater than critical t- values, hence we reject the null hypothesis. Hence $\mu_{MGT} > \mu_{LDA}, \mu_{LSA}, \mu_{BERTopic}, \mu_{Top2Vec}$

## 7 Conclusion

Hence it can be interpreted that MGT topic model shows improvement in semantic coherence of topics compared to topic modelling using LDA. Hence this model can be useful in identifying latent themes from unstructured text with greater interpretability. Moreover, this model incorporates the relationship between words in the form of co-variance matrix, thus ensuring that related words get clubbed in a topic, thus improving semantic coherence and topic interpretability. Thirdly, analysis of topic assignments show that that sentences and paragraphs are assigned same topic, thus bridging another drawback of LDA models which may assign multiple topics to words even in a single sentence. Finally, this model assigns single unique topic to short documents instead of occasional allocation of multiple topics to short documents by LDA. This is consistent with the fact that a sentence, and even a paragraph, represents a single topic in a document.

Despite the advantages, the model suffers from a few drawbacks. One drawback of the model is that Bag-of-Words approach is used for identifying topics, whereas word sequencing is not considered. Secondly, the model needs to be tried on bigger datasets and bigger documents (i.e. Wikipedia datasets) to see how they perform and the computational time taken to identify topics.

**Disclosure of Interests.** The authors declare that they have no known competing financial interests or personal relationships that could have appeared to influence the work reported in this paper.